\documentclass[journal]{IEEEtran}

\ifCLASSOPTIONcompsoc
  \usepackage[nocompress]{cite}
\else
  \usepackage{cite}
\fi

\usepackage{graphicx}
\usepackage{subfigure}
\usepackage{multirow}
\usepackage{amssymb}
\usepackage{amsmath}
\usepackage{booktabs}
\usepackage[ruled]{algorithm2e}
\usepackage{color}
\usepackage{ragged2e}
\usepackage{hyperref}
\usepackage{cite}
\usepackage{array}
\usepackage{colortbl}

\ifCLASSINFOpdf
\else
\fi

\begin{document}
%

\title{Carrying out CNN Channel Pruning in a White Box}

\author{Yuxin Zhang,
        Mingbao Lin,
        Chia-Wen Lin,~\IEEEmembership{Fellow,~IEEE}, 
        Jie Chen,~\IEEEmembership{Member,~IEEE}, \\
        Yongjian Wu, 
        Yonghong Tian,~\IEEEmembership{Senior Member,~IEEE},
        and Rongrong Ji,~\IEEEmembership{Senior Member,~IEEE}
\IEEEcompsocitemizethanks{\IEEEcompsocthanksitem Y. Zhang is with the Media Analytics and Computing Laboratory, Department of Artificial Intelligence, School of Informatics, Xiamen University, Xiamen 361005, China.
\IEEEcompsocthanksitem M. Lin is with the Media Analytics and Computing Laboratory, Department of Artificial Intelligence, School of Informatics, Xiamen University, Xiamen 361005, China, also with the Youtu Laboratory, Tencent, Shanghai 200233, China.
\IEEEcompsocthanksitem C.-W. Lin is with the Department of Electrical Engineering and the Institute of Communications Engineering, National Tsing Hua University, Hsinchu 30013, Taiwan.
\IEEEcompsocthanksitem J. Chen is with the Peng Cheng Laboratory, Shenzhen 518000, China.
\IEEEcompsocthanksitem Y. Wu is with the Youtu Laboratory, Tencent, Shanghai 200233, China.
\IEEEcompsocthanksitem Y. Tian is with Department of Computer Science and Technology, Peking University, Beijing 100871, China, also with  the Peng Cheng Laboratory, Shenzhen 518000, China.
\IEEEcompsocthanksitem R. Ji (Corresponding  Author) is with the Media Analytics and Computing Laboratory, Department of Artificial Intelligence, School of Informatics, Xiamen University, Xiamen 361005, China, also with Institute of Artificial Intelligence, Xiamen University, Xiamen 361005, China, and also with the Peng Cheng Laboratory, Shenzhen 518000, China (e-mail: rrji@xmu.edu.cn). 
}
\thanks{Manuscript received April 19, 2005; revised August 26, 2015.}}

\markboth{IEEE TRANSACTIONS ON NEURAL NETWORKS AND LEARNING SYSTEMS}%
{Shell \MakeLowercase{\textit{et al.}}: Bare Demo of IEEEtran.cls for IEEE Journals}

\maketitle

\begin{abstract}
Channel Pruning has been long studied to compress CNNs, which significantly reduces the overall computation. Prior works implement channel pruning in an unexplainable manner, which tends to reduce the final classification errors while failing to consider the internal influence of each channel. In this paper, we conduct channel pruning in a white box. Through deep visualization of feature maps activated by different channels, we observe that different channels have a varying contribution to different categories in image classification. Inspired by this, we choose to preserve channels contributing to most categories. Specifically, to model the contribution of each channel to differentiating categories, we develop a class-wise mask for each channel, implemented in a dynamic training manner w.r.t. the input image's category. On the basis of the learned class-wise mask, we perform a global voting mechanism to remove channels with less category discrimination. Lastly, a fine-tuning process is conducted to recover the performance of the pruned model. To our best knowledge, it is the first time that CNN interpretability theory is considered to guide channel pruning. Extensive experiments on representative image classification tasks demonstrate the superiority of our White-Box over many state-of-the-arts. For instance, on CIFAR-10, it reduces 65.23\% FLOPs with even 0.62\% accuracy improvement for ResNet-110. On ILSVRC-2012, White-Box achieves a 45.6\% FLOPs reduction with only a small loss of 0.83\% in the top-1 accuracy for ResNet-50. Code is available at \url{https://github.com/zyxxmu/White-Box}.

\end{abstract}


\maketitle


\IEEEpeerreviewmaketitle

\section{Introduction}
\label{introduction}
\IEEEPARstart{T}{hough} convolutional neural networks (CNNs) have shown predominant performance in image classification tasks~\cite{simonyan2015very, he2016deep}, the vast demand on computation cost has prohibited them from being deployed on edge devices such as smartphones and embedded sensors. To address this, the researchers have developed several techniques for CNNs compression, such as network pruning~\cite{han2015learning,ding2019global}, parameter quantization~\cite{hubara2016binarized, liu2020bi}, tensor decomposition~\cite{peng2018extreme, hayashi2019exploring} and knowledge distillation~\cite{romero2014fitnets,hinton2015distilling}, \emph{etc}. Among them, channel pruning has attracted ever increasing attention for its easy combination with general hardware and  Basic Linear Algebra Subprograms (BLAS) libraries, which is thus the focus of this paper.

channel pruning removes the entire channels to generate a sub-network of the original CNN with less computation cost. Existing studies could roughly be categorized into three categories. The first group abides a three-step pruning pipeline including pre-training a dense model, selection of ``important'' filters and fine-tuning the sub-net. Typically, most works of this category focus on the second step by either figuring out a filter importance estimation, such as $\ell_1$-norm~\cite{li2017pruning}, geometric information~\cite{he2019filter} and activation sparsity~\cite{hu2016network}, or regarding channel pruning as an optimization problem~\cite{lin2020hrank, guo2020channel}. The second category implements channel pruning with additional sparsity constraints~\cite{liu2017learning, luo2020autopruner, ding2019centripetal, ding2020lossless}, after which, the pruned model can be available by removing zeroed channels or channels below a given threshold. The last group applies AutoML techniques to directly search the channel number of each layer under a given computation budget, thus achieve pruning in an automatic manner~\cite{he2018amc, yang2018netadapt, liu2019metapruning, liu2021joint}.

\begin{figure*}[h]
\begin{center}
\includegraphics[width=0.85\linewidth,height=0.2\pdfpageheight]{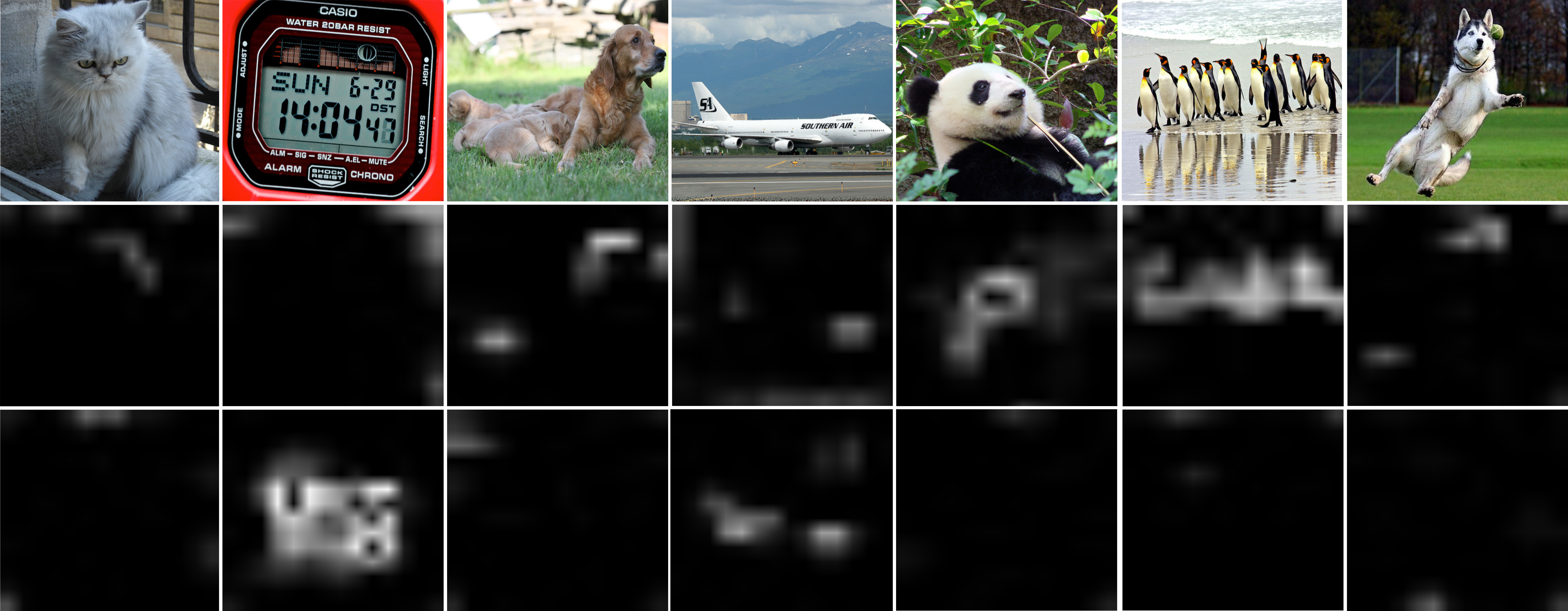}
\end{center}
\vspace{-1.0em}
\caption{\label{visualization}Visualization of images from different categories (first row) along with the feature maps of the 5-th and 144-th channels (second and third row, respectively) in the conv-12 layer of VGG16-Net~\cite{simonyan2015very} trained on ImageNet.  
As can be seen, with different scene, light, pose of images as input, head and textual information are always activated by these two channels respectively, despite that the ImageNet dataset does not contain such categories.
All feature maps are with their original activation value and size.
\vspace{-1.5em}
}
\end{figure*}

Despite the progress, existing methods build channel pruning by observing the CNN output, \emph{i.e.}, the final classification performance, while leaving the internal influence of a CNN model hardly touched. For example, Li \emph{et al}.~\cite{li2017pruning} removed filters with smaller $\ell_1$-norm, which can indeed be viewed as to minimize the output difference between the original model and the pruned model. To take a more in-depth analysis, the massive non-linear operations inside CNNs make them hardly understandable. Thus, existing methods~\cite{li2017pruning, ding2019approximated} choose to regard the CNNs as a black box and observe the final output for network pruning. 
For instance, Ding \emph{et al}.~\cite{ding2019approximated} leveraged binary search to remove filters with the least accumulated errors calculated by the final output.
From this perspective, we term these methods ``Black-Box pruning'' in this paper.

Nevertheless, understanding the internal explanation of deep CNNs has attracted increasing attention~\cite{wu2017interpretable,yosinski2015understanding,zeiler2014visualizing,zhang2018interpretable,zhou2014object}, which also advances various vision tasks. For instance, Zeiler \emph{et al.}~\cite{zeiler2014visualizing} won the championship of the ILSVRC-2013 by adjusting architecture through visualization of internal feature maps. Inspired by this, we believe that exploring the internal logic in CNNs could be a promising prospect to guide channel pruning.

As exploited in~\cite{yosinski2015understanding}, the feature maps of each channel have the locality that a particular area in one feature map is activated. Inspired by this, we visualize the feature maps generated by VGG16-Net~\cite{simonyan2015very} trained on ImageNet to explore the local information in the internal layers of CNNs. As can be seen from Fig.\,\ref{visualization}, the 5-th channel at the 12-th convolutional layer always generates feature maps that contain head information while the 144-th channel attempts to activate textual information. Even though there is no explicitly labeled head or text, this CNN model automatically learns to extract partial information to make better decisions, which exactly meets human intuition when classifying an image. That is, head information extracted by the 5-th channel helps the network to identify animals, and textual information extracted by the 144-th channel contributes to classify categories with texts such as digital watches. However, some local features may not be beneficial to identifying all categories. For example, the 144-th channel always chooses to deactivate most of the pixels when processing images with no textual semantics like dogs and pandas (see the third and fifth columns in Fig.\,\ref{visualization}). Such local representation on the internal layers of a CNN shows that channels have varying contribution to different categories in image classification, which motivates us to rethink the importance criterion of channel pruning.

Instead of simply considering the CNN output after removing a channel as prior arts do, we target at finding each channel's contribution to identifying different kinds of images. It is intuitive that if feature maps activated by one channel can benefit most categories' classification, this channel is essential and should be preserved; otherwise, it can be safely removed.

To this end, we assign each channel a class-wise mask, the length of which is basically the same as the category number in the training set. Specifically, we utilize the ground-truth labels as auxiliary information in pruning the network. For each category of the input images, the corresponding mask is activated to multiply on the output feature map for model inference. If one channel generates feature maps that have a positive recognition effect on some categories, masks corresponding to these categories will receive large training loss gradients. By exerting a sparsity constraint that pulls the class-wise mask toward zero to counteract such gradients, these masks will maintain relatively large absolute values. On the contrary, if this channel contributes little to most categories, then the corresponding masks will be punished close to zero. Thus, after a few training epochs, each channel's importance score can be measured by the absolute sum of its class-wise mask, reflecting its overall contribution to identifying all categories. In this way, we can carry out the pruning in an explainable manner, for which we term our pruning as ``White-Box".

We further propose an iteratively global voting, which is performed using the above importance score to remove unimportant channels until the FLOPs of the pruned model meet the pre-given computation budget. It is worth mentioning that the layer-wise pruning rate can be decided in an automatic manner, which demonstrates the efficiency of White-Box compared with the previous works using hand-crafted designs~\cite{li2017pruning,lin2020hrank} or conducting time-consuming search~\cite{he2018amc,liu2019metapruning}. Lastly, a fine-tuning process is conducted to recover the performance of the pruned network.

\begin{figure*}[!t]
\begin{center}
\includegraphics[width=0.8\linewidth,height=0.17\pdfpageheight]{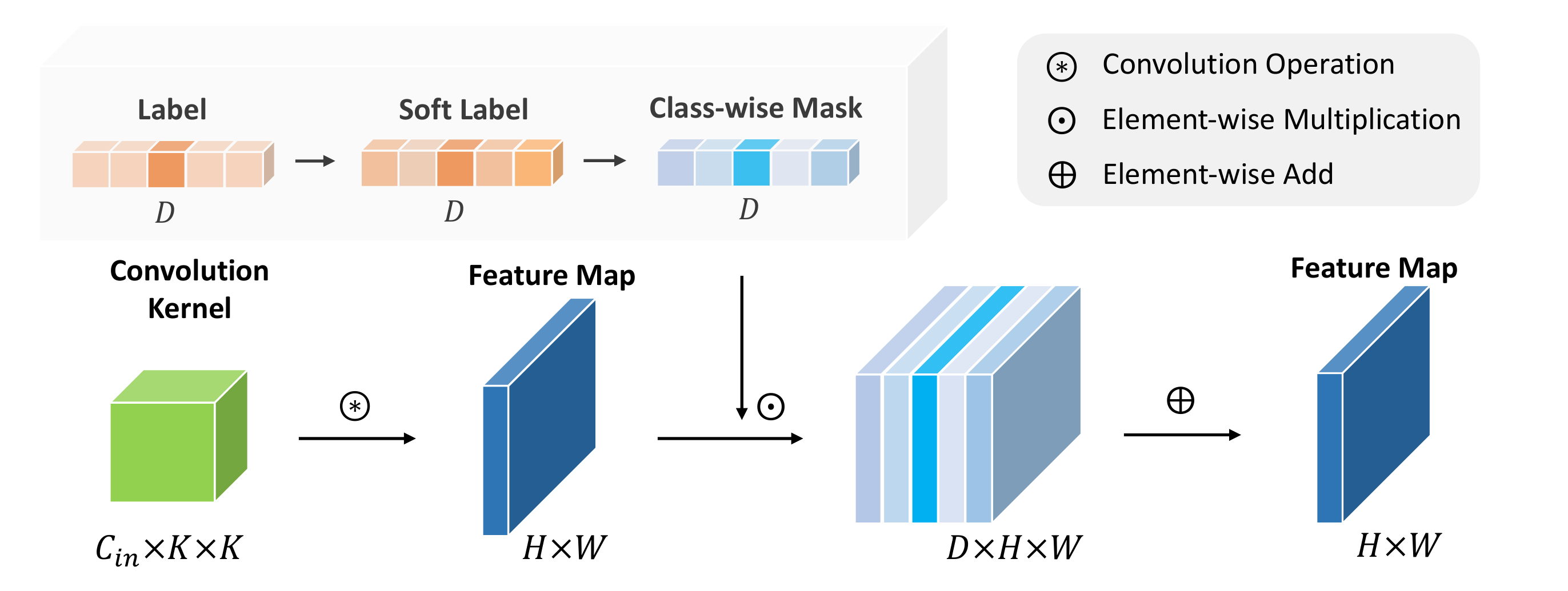}
\end{center}
\setlength{\abovecaptionskip}{0pt}
\setlength{\belowcaptionskip}{0pt}
\vspace{-1.0em}
\caption{\label{framework}Framework of the proposed White-Box for class-wise mask training. For better representation, we choose one channel for description. Our White-Box will assign a class-wise mask for this channel with a dimension of category number $D$ in the dataset, which aims at measuring the contribution of this channel to different categories' identifying. With a specific category of the input image, we first soft the one-hot vector of the ground-truth label to activate the class-wise mask for stable training. Output feature map of this channel will then be multiplied by the class-wise mask, and be added in an element-wise manner to generate the final feature map for model inference. By doing so, the class-wise mask will receive gradient signals respecting to input images' category. Therefore, White-Box will be able to find this channel's contribution to recognizing different categories. (Best viewed with zooming in)
\vspace{-1.5em}
}
\end{figure*}

Our contributions are summarized as follows:
\begin{itemize}
\item Based on an in-depth analysis of CNNs interpretation, we propose a novel explainable importance criterion for channel pruning that we should preserve channels beneficial to identifying most categories. To our best knowledge, this is the first time that CNN interpretability theory is considered to guide channel pruning. 
\item We carry out channel pruning in a white box by jointly training a class-wise mask along with the original network to find each channel's contribution for classifying different categories. Then a global voting and a fine-tuning are conducted to obtain the final pruned model.
\item Extensive experiments on CIFAR-10 and ILSVRC-2012 demonstrate the advantages of the proposed White-Box over several state-of-the-art advances in accelerating the CNNs for image classification. 
\end{itemize}

\section{Related Work}

\textbf{Channel Pruning.}
Channel pruning targets at snipping away entire channels in convolution kernel to obtain a pruned model, which not only saves computation cost, but is also compatible with off-the-shelf hardware. As discussed in Sec.\,\ref{introduction}, previous channel pruning works can be approximately divided into three groups. 
Starting from a pre-trained model, the first category designs various importance criteria to remove unimportant channels. For example, Li \emph{et al}.~\cite{li2017pruning} chose to prune filters with smaller $\ell_1$-norm. SASL~\cite{shi2020sasl} proposed to measure the importance of channels using both the prediction performance and computation consumption. He \emph{et al}.~\cite{he2017channel} considered the construction error of the next layer as an importance criterion and conducted pruning in a layer-by-layer fashion. Guo \emph{et al}.~\cite{guo2020model} further proposed to iteratively remove a group of channels from several selected layers instead of a single layer. Nevertheless, Guo \emph{et al}.~\cite{guo2020channel} observed the ``next-layer feature map removal'' problem that if a feature from the next layer will be removed at the next pruning stage, minimizing the reconstruction error of such features is unnecessary. To solve this, they considered both classification loss and feature importance as a pruning criterion to deal with the influence of removing next-layer feature maps. 
There are also some approaches that judge the importance of channels using attention modules~\cite{yu2020antidote,yamamoto2018pcas,wang2019convolutional}. For example, Wang \emph{et al.}~\cite{wang2019convolutional} leveraged the attention module to obtain the scaling factors of each channels, which are then served as the importance criterion. Our approach differs from attention-based pruning methods in that it considers the class-wise contribution of a single channel on the basis of CNN interpretability, while attention based methods focus on the channel-wise attention scores.
The second group implements channel pruning in a training-adaptive manner by introducing extra sparsity regularization. For example, Huang \emph{et al}.~\cite{huang2018data} introduced a scaling factor to scale the outputs of specific structures and added sparsity on these factors. They then trained the sparsity-regularized mask for network pruning through data-driven selection. Luo \emph{et al}.~\cite{luo2020autopruner} employed an ``autopruner'' layer appended in the convolutional layer to prune filters automatically. By regularizing auxiliary parameters instead of original weights values, Xiao \emph{et al}.~\cite{xiao2019autoprune} pruned the CNN model via a gradient-based updating rule. Chen \emph{et al}.~\cite{chen2020dynamical} designed a channel-wise gate to dynamically estimate the conditional accuracy change and gradually prune channels during training process. 
Tang \emph{et al}.~\cite{tang2021manifold} further explored the manifold information to dynamically excavate the channel redundancy of CNNs.
The last category of channel pruning approaches apply AutoML techniques to directly search the channel number of each layer under a given computation budget~\cite{he2018amc, liu2019metapruning}. For instance, He \emph{et al}.~\cite{he2018amc} proposed to automate the searching process by reinforcement learning. Liu \emph{et al}.~\cite{liu2019metapruning} trained a PruningNet in advance to predict the weights of candidate networks and leverage evolutionary algorithm to search for the bset candidate. Unfortunately, all of these methods conduct channel pruning with respect to the CNN output, failing to consider the internal mechanism of a CNN model. Though there are considerable improvements, interpretation for channel pruning remains an open problem.

\textbf{CNN Interpretation.}
Despite an impressive performance in various tasks, CNNs have long been known as ``Black-Box'' for its end-to-end learning strategy. As an Achilles' heel, CNN interpretability has attracted increasing attention in recent years~\cite{wu2017interpretable,yosinski2015understanding,zeiler2014visualizing,zhang2018interpretable,zhou2014object}. Most studies of understanding CNN representations fall into visualization. Zeiler \emph{et al}.~\cite{zeiler2014visualizing} proposed a visualizing technique by projecting the feature activations back to the input pixel space to observe the function of intermediate feature maps. Yosinkin \emph{et al}.~\cite{yosinski2015understanding} further developed a tool that visualizes the activations produced on each layer of a trained CNN model. They then gleaned several surprising intuitions using this tool, including that some channels always represent useful partial information for classification decisions like faces or text, although there are no explicit labels for these items. Such phenomenon is also found in~\cite{zhou2014object}, which motivates us to consider what kind of channels should be pruned from an interpretable perspective. Recently, CNN interpretation theory also demonstrates its effectiveness by achieving state-of-the-art results on various tasks such as image classification~\cite{zhang2018interpretable} and object detection~\cite{wu2017interpretable}. Hence we believe CNN interpretability could be a superior foreground for guiding channel pruning.

\section{Proposed Method}
\subsection{Background}\label{preliminary}
Considering an $L$-layer CNN model, its kernel weights can be represented as $\mathcal{W} = \{\mathcal{W}^1, \mathcal{W}^2, ..., \mathcal{W}^L\}$. The kernel in the $l$-th layer is denoted as $\mathcal{W}^l \in \mathbb{R}^{C^l_{out} \times C^l_{in} \times K^l \times K^l}$, where $C^l_{out} $, $C^l_{in}$, $K^l$ denote the numbers of output channels and input channels, and the kernel size, respectively. Let $ \mathcal{I}^{l} \in \mathbb{R}^{N \times C_{in}^l \times H^l \times W^l} $ be the input of the $l$-th layer where $N$ is the batch size of input images, and $H^l, W^l$ respectively stand for the height and width of the input. Given a batch of training image set $X$, associated with a set of class labels $Y^{N \times D}$ where $D$ represents the total number of categories in the training set, we denote $X = \mathcal{I}^1$. For the $i$-th input image $X_{i, :, :, :}$, we treat its label $Y_{i, :}$ as a one-hot vector. Sometimes we simply denote $Y_{i, :} = j$ to indicate that the $j$-th entry is set to one while the others are zero\footnote{It also indicates that image $X_{i, :, :, :}$ belongs to the $j$-th class.}. With a conventional CNN, the output of the $l$-th layer can be calculated as \footnote{The convolutional operation usually involves a bias term and is followed by a non-linear operation. For ease of representation, we omit them in this paper.}
\begin{equation}\label{conv}
\mathcal{O}^l = \mathcal{I}^l \circledast \mathcal{W}^l,
\end{equation}
where $\circledast$ denotes the convolutional operation. We have $\mathcal{O}^l = \mathcal{I}^{l+1} \in \mathbb{R}^{N \times C^{l+1}_{in} \times H^{l+1} \times W^{l+1}}$ and $C^{l}_{out} = C^{l + 1}_{in}$. Then, its training loss can be expressed as
\begin{equation}\label{original_loss}
\min_{\mathcal{W}} \; \mathcal{L}(X,\mathcal{W};Y).
\end{equation}

For channel pruning, a subgroup of output channels in $\mathcal{W}^l$ will be removed to obtain pruned kernel $\widehat{\mathcal{W}}^{l} \in \mathbb{R}^{\widehat{C}^l_{out} \times \widehat{C}^l_{in} \times K^l \times K^l}$ under the constraints of $\widehat{C}^l_{out} \le C^l_{out}$ and $\widehat{C}^l_{in} \le C^l_{in}$, thus it can reach a better trade-off between computation cost and accuracy performance. It is worth mentioning that the corresponding input channels of $\mathcal{W}^{l+1}$ are also removed. Accordingly, we can reformulate Eq.\,(\ref{conv}) and Eq.\,(\ref{original_loss}) in the pruned network as follows:
\begin{equation}\label{pruned_conv}
\widehat{\mathcal{O}}^l = \widehat{\mathcal{I}}^l \circledast \widehat{\mathcal{W}}^l,
\end{equation}
\begin{equation}\label{pruned_loss}
\min_{\widehat{{\mathcal{W}}}} \; \mathcal{L}(X,\widehat{\mathcal{W}};Y).
\end{equation}

As discussed in Sec.\,\ref{introduction}, most prior works perform channel pruning by directly judging channels' redundancy using an importance estimation or imposing sparsity penalty to dynamically conduct channel pruning. Essentially, these methods complete network pruning based on the final network outputs in an unexplainable way, which neglects the internal influence of channels, thus we term these methods ``Black-Box pruning.''

In contrast, for the first time, we conduct channel pruning by exploring the internal influence of CNNs. Our motivation is based on the observation in~\cite{yosinski2015understanding}, which reveals that the representations in the internal CNN layers are surprisingly local, implying that many channels are only responsible for extracting partial information. We argue that some of these partial information is redundant and may not be beneficial to classify all categories, as illustrated in Fig.\,\ref{visualization}. Hence, it is crucial to identify each channel's ability to derive recognizable local features that contribute to recognizing categories.

\subsection{Pipeline of White-Box}
In order to address the above issues, we propose a novel explainable channel pruning method, termed ``White-Box.'' 
The purpose of White-Box is to find channels that can generate feature maps containing discriminative category information as much as possible, so as to retain those that contribute to the recognition of most categories, and prune those channels that only benefit few categories.
Specifically, we firstly design a class-wise mask to multiply on each channel to guide a class-wise training.
When training a particular class of images, the corresponding mask will be activated for model inference and backpropagation, which will be described in detail in Sec.\,\ref{CMask}.
Fig.\,\ref{framework} shows the framework of White-Box for class-wise mask training. Subsequently,  as explained in Sec.\,\ref{IGV},  we propose a global voting mechanism to preserve those channels that make contributions to the recognition of most classes, as well as to automatically determine the layer-wise pruning rate without manual involvement.
Finally, a fine-tuning process is conducted to boost the performance of the pruned model.
Our White-Box is summarized in Alg.\,\ref{alg1}.
\subsection{Class-wise Mask}\label{CMask}

\begin{algorithm}[t]
\caption{\label{alg1}Algorithm Description of White-Box}

\LinesNumbered
\KwIn{An L-layer CNN, global pruning rate $\alpha$, mask training epochs $\mathcal{T}$}
\KwOut{The pruned model and its parameters $\widehat{\mathcal{W}}$}
Initialize model parameters $\mathcal{W}$ , pruning rate $\widehat{\alpha} = 0$, class-wise mask $\mathcal{M} =  \{1\}$\;
\For{$h = 1 \rightarrow \mathcal{T}$}
{
	Train model with $\mathcal{W}$ and $\mathcal{M}$ via Eq.\,\ref{objectfunc}\;
}
Get criterion score $\mathcal{S}$ via Eq.\,\ref{score}\;
Sort S in ascending order\;
\While{$\widehat{\alpha} \leq \alpha$}
{
	Remove the first element in $\mathcal{S}$ and its corresponding channel in $ \mathcal{W}$ and mask in $\mathcal{M}$\;
	Update current FLOPs pruning rate $\widehat{\alpha}$\;
}
Integrate $\widehat{\mathcal{M}}$ into $\widehat{\mathcal{W}}$ via Eq.\,\ref{integrate} \;
Fine-tune pruned model to recover performance.
\end{algorithm}

The core of our White-Box is to assign per-layer kernel $\mathcal{W}^l$ a class-wise mask, which is formatted in the form of $\mathcal{M}^l \in \mathbb{R}^{D \times C_{out}}$. Specifically, the mask value $\mathcal{M}^l_{j, c}$ is built to measure the contributions of individual channels $\mathcal{W}^l_{c, :, :, :}$ to the network for recognizing the $j$-th category.

Then, for the $i$-th input image $X_{i, :, :, :}$ with label $Y_{i, :}$, the convolution using Eq.\,(\ref{conv}) in the forward propagation under our mask framework can be rewritten as
\begin{equation}\label{conv2}
\begin{split}
\mathcal{O}^l_{i,:,:,:} &=  \mathcal{I}^l_{i,:,:,:} \circledast (  \mathcal{M}^l_{Y_{i, :},:} \ast  \mathcal{W}^l), \;\; i = 1, 2,..., N,
\end{split}
\end{equation}
where $\ast$ denotes the channel-wise multiplication, that is, channel $\mathcal{W}^l_{c, :, :, :}$ is multiplied with the scalar mask $\mathcal{M}^l_{j, c}$. Subsequently, our training loss can be obtained as:
\begin{equation}\label{pruned_loss}
\min_{{\mathcal{W}}, \mathcal{M}} \; \mathcal{L}(X,\mathcal{W}, 
\mathcal{M};Y).
\end{equation}

The rationale of our mask design lies in that, during backpropagation, the mask $\mathcal{M}^l_{j, c}$ will receive the gradient signals regarding the input images of the $j$-th category. On the premise of this principle, if channel $\mathcal{W}^l_{c, :, :, :}$ benefits the network to recognize input images from the $j$-th category, $\mathcal{M}^l_{j, c}$ will be positively activated, and deactivated, otherwise. Therefore, our class-wise mask design can well reflect the internal logic in CNNs, which seamlessly follows our motivation behind the channel pruning in our white box.
In comparison with typical CNNs where the label information is utilized in the loss layer, our class-wise mask-based convolutional operations are more label-guided since it requires label information in every convolutional layer as shown in Eq.\,(\ref{conv2}). This poses a critical challenge of the over-fitting problem since the label information is in the format of one-hot vector, meaning that we need to provide ground-truth labels for each convolutional layer’s forward propagation. Such data flow during training varies largely from the real testing part, thus may cause the over-fitting problem.

Inspired by the label-smoothing regularization~\cite{szegedy2016rethinking},
we propose to solve this problem by softening the one-hot vector, denoted as $\widehat{Y} \in \mathbb{R}^{N \times D}$, element of which is defined as
\begin{equation}
    \widehat{Y}_{i, d} = 
    \begin{cases}
    Y_{i, d}, & \mbox{if $Y_{i, d}$ = 1}, \\
    \mathcal{N}(0,1),   & \mbox{otherwise},
    \end{cases}
\end{equation}
where $\mathcal{N}(\cdot,\cdot)$ denotes the normal distribution.

Then, the convolution in Eq.\,(\ref{conv2}) is reformulated as
\begin{equation}\label{conv_soft}
\begin{split}
\mathcal{O}^l_{i,:,:,:} = \sum_{d=1}^D \mathcal{I}^l_{i,:,:,:} &\circledast \Big( \big( \widehat{Y}_{i,d} \cdot  \mathcal{M}^l_{d, :}\big) \ast  \mathcal{W}^l \Big), \\&i = 1, 2, ..., N.
\end{split}
\end{equation}    

Thus, channel pruning can be realized by removing those channels with poor masks. It is natural to impose sparsity constraint on per-channel mask as
\begin{equation}\label{mask_sparsity}
    \min_{\mathcal{M}}\sum_{l=1}^L\sum_{c=1}^{C^l_{out}}\| \mathcal{M}^l_{:, c} \|_2.
\end{equation}

Note that we choose $\ell_2$-norm instead of $\ell_1$-norm as previous sparsity regularization works do~\cite{luo2020autopruner,xiao2019autoprune}. The rationale falls in that our object is not to regularize the masks to exactly 0s that $\ell_1$ norm leads to, but to measure the class-wise contribution of each channel. After training, we leverage global voting to directly obtain the pruned model, which will be introduced in the next section. Thus, we choose to leverage $\ell_2$-norm for its smoothness and rotation invariance.

Eq.\,(\ref{pruned_loss}) and Eq.\,(\ref{mask_sparsity}) lead to our final training loss:
\begin{equation}\label{objectfunc}
  \min_{\mathcal{W}, \mathcal{M}} \mathcal{L}(X, \mathcal{W}, \mathcal{M}; Y) + \lambda \sum_{l=1}^L\sum_{c=1}^{C^l_{out}}\| \mathcal{M}^l_{:, c} \|_2.
\end{equation}

Noticeably, the objective of Eq.\,(\ref{objectfunc}) targets at locating channels that contribute more to recognizing the input images, which then make up of the pruned kernel $\widehat{W}$ as described in Sec.\,\ref{preliminary}, followed by a series of fine-tuning procedures using loss objective of Eq.\,(\ref{pruned_loss}). Therefore, only a few epochs are needed to train our class-wise mask so as to derive $\widehat{\mathcal{W}}$ in our empirical observation\footnote{We consider 10\% of the total fine-tuning epochs for training the class-wise mask.}.

\subsection{Global Voting for Cross-layer Pruning}\label{IGV}

Given a global pruning rate $\alpha$, how to appropriately distribute it to each layer to preserve $C^l_{out}$ channels would significantly affect the performance of the pruned model~\cite{liu2019rethinking}. Prevalent methods resort to rule-of-thumb designs~\cite{li2017pruning, lin2020hrank} or complex structure search~\cite{liu2019metapruning, li2020eagleeye}.
Fortunately, our White-Box can tacitly obtain a global important criterion for all channels in the network and conduct layer-wise pruning rate decision in an iteratively-voting manner. Detailedly, considering a trained class-wise mask $ M^l_{:,c} \in \mathbb{R}^D$ of the $c$-th channel in the $l$-th layer, each item in this tensor represents this channel's ability for classifying one corresponding category of the dataset, thus we can measure this channel's contribution to overall classification performance by simply summing up these class-wise mask scores. We denote all scores of $\mathcal{M}^l$ as $\mathcal{S}^l \in  \mathbb{R}^{C^l_{out}}$:
\begin{equation}\label{score}
 \mathcal{S}_c^l = \sum_{d=1}^D M^l_{d,c},\; c = 1\;...\;C^l_{out},
\end{equation}
which then will serve as importance criterion for this channel. 
Given a global pruning rate $\alpha$,  after obtaining all channels' importance scores $\mathcal{S}$ in the whole network, we iteratively remove the least-impact channels and calculate FLOPs pruning rate $\widehat{\alpha}$ of the current model until $\widehat{\alpha} \geq \alpha$. After voting, we integrate the left class-wise mask $\widehat{\mathcal{M}}$ into $\widehat{\mathcal{W}}$ to conduct fine-tuning for performance recovery. Particularly, as we soften the label obeying a standard normal distribution $\mathcal{N}(\mu = 0.5, \sigma = 1) $ during training except for the ground-truth related current input, the overall pruned $\widehat{\mathcal{M}}$ can be mixed into $\widehat{\mathcal{W}}$ by
\begin{equation}\label{integrate}
 \widehat{\mathcal{W}} =  \; \widehat{\mathcal{W}} \ast \sum_{d=1}^D \mu \;\widehat{M_{d,:}}.
\end{equation}

Lastly, more epochs are used to fine-tune the pruned model.
%

\section{Experiments}
\subsection{Implementation Details}
\textbf{Datasets and Backbones.} We conduct extensive experiments on two representative datasets including CIFAR-10~\cite{krizhevsky2009learning} and ILSVRC-2012~\cite{russakovsky2015imagenet} to demonstrate the efficacy of the proposed White-Box. We prune prevailing CNN models including VGG-16~\cite{simonyan2015very}, ResNet-56/110~\cite{he2016deep}, MobileNet-v2~\cite{sandler2018mobilenetv2} on CIFAR-10 and ResNet-50~\cite{he2016deep} on ILSVRC-2012.

\begin{figure*}[!t]
\begin{center}
\includegraphics[width=0.9\linewidth,height=0.18\pdfpageheight]{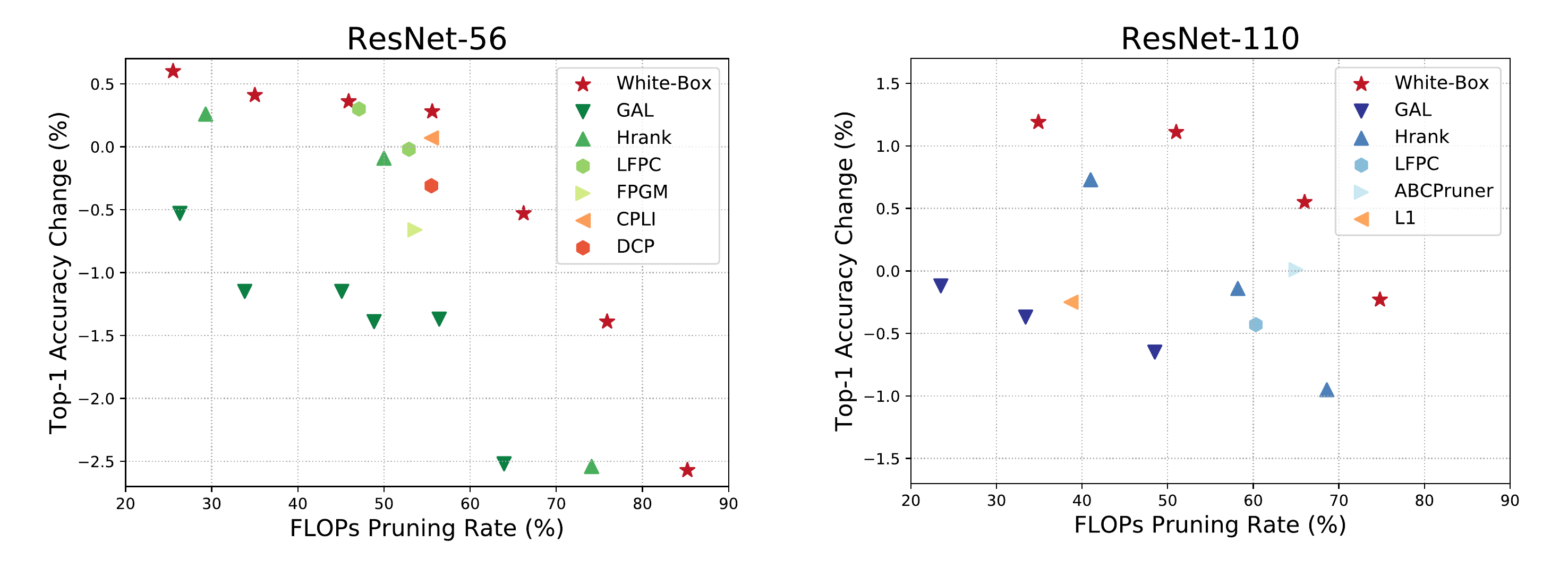}
\end{center}
\setlength{\abovecaptionskip}{0pt}
\setlength{\belowcaptionskip}{0pt}
\vspace{-1.4em}
\caption{\label{vary_rate}Top-1 accuracy comparison between existing methods and the proposed White-Box under different pruning rates of FLOPs. The experiments are conducted using ResNet-56 and ResNet-110 on CIFAR-10. 
\vspace{-1.8em}
}
\end{figure*}

\textbf{Configurations.} 
We set the sparse parameter $\lambda$ as $10^{-2}$ for VGGNet-16 and MobileNet-v2, and $5 \times 10^{-4}$ for ResNets. Then, We train our class-wise masks using the original full network with a learning rate of 0.1 for 30 epochs on CIFAR-10 and 9 epochs on ILSVRC-2012. After the global voting, the pruned model is then fine-tuned via the SGD optimizer. The momentum, batch size are set to 0.9, 256, respectively, in all experiments. 
On CIFAR-10, we iterate 300 epochs to fine-tune the pruned model \ with an initial learning rate of 0.1, which is divided by 10 at the 150-th and 225-th epochs. 
On ILSVRC-2012, ResNet-50 is fine-tuned for 90 epochs with step scheduler learning rate, which begins at 0.1 and is divided by 10 every 30 epochs. 
The weight decay rate is set to $5 \times 10^{-4}$ for all models except for MobileNet-v2: $4 \times 10^{-5}$  on CIFAR-10 and $10^{-4}$ on ILSVRC-2012. 
All experiments are implemented with Pytorch~\cite{pytorch2015} and run on NVIDIA Tesla V100 GPUs.
The data argumentation includes crop and horizontal flip.

\begin{table}[!t]
\centering
\small
\caption{Results for pruning VGGNet-16 on CIFAR-10.}
\vspace{-0.5em}
\label{vgg_cifar10}
\setlength{\tabcolsep}{0.6em}
\begin{tabular}{cccc}
\toprule
 Model           &Top-1 Acc.& Acc.~$\downarrow$        & FLOPs $\downarrow$  \\

\midrule
$\ell_1$~\cite{li2017pruning}           &93.25\% $\rightarrow$ 93.40\%     & -0.25\%  &34.2\%   \\
GAL~\cite{lin2019towards}   &93.02\% $\rightarrow$ 92.03\%    &1.93\%       &39.6\%     \\
SSS~\cite{huang2018data}            &93.02\% $\rightarrow$ 93.02\%   &0.00\%         &41.6\%     \\ 
Slimming~\cite{liu2017learning} & 93.66\% $\rightarrow$ 93.80\% &-0.14\% & 51.0\% \\

HRank~\cite{lin2020hrank}     &93.02\% $\rightarrow$ 91.23\%  &1.79\%         &76.5\%  \\

\textbf{White-Box}           &~93.02\% $\rightarrow$ \textbf{93.47\%}& \textbf{-0.45\%}     &\textbf{76.4\%} \\

\bottomrule
\end{tabular}
\end{table}

%
\begin{table}[!t]
\setlength{\tabcolsep}{0.6em}
\centering
\small
\caption{Results for pruning ResNet-56 on CIFAR-10.}
\vspace{-0.5em}
\label{resnet56_cifar10}
\begin{tabular}{ccccc}
\toprule
Model           &Top-1 Acc.& Acc.~$\downarrow$       & FLOPs   $\downarrow$  \\
\midrule
HRank~\cite{lin2020hrank}             &93.26\% $\rightarrow$ 93.17\%     &0.09\%       &50.0\%     \\
AMC~\cite{he2018amc} & 92.80\% $\rightarrow$ 91.90\% &0.90\% &50.0\%  \\
SCP~\cite{kang2020operation} &93.69\% $\rightarrow$ 93.23\% &0.46\%  &51.5\%  \\
SFP~\cite{he2018soft} &93.59\% $\rightarrow$ 92.26\% &1.33\%  &52.6\%  \\
LFPC~\cite{he2020learning} &93.26\% $\rightarrow$ 93.24\% &0.02\%  &52.9\% \\
DSA~\cite{ning2020dsa} &93.12\% $\rightarrow$ 92.91\% & 0.21\% & 53.2\% \\
FPGM~\cite{he2019filter} & 93.59\% $\rightarrow$ 92.93\% &0.66\%&  53.6\% \\
\textbf{White-Box}                              &~93.26\% $\rightarrow$ \textbf{93.54\%}&\textbf{-0.28\%}& \textbf{55.6\%}     \\

\bottomrule
\end{tabular}

\end{table}

%
\subsection{Comparison on CIFAR-10}
We first demonstrate the superiority of White-Box on CIFAR-10. The FLOPs pruning rate of the compressed models and their top-1 accuracy performance are reported. The accuracy reported is in the format of ``pre-trained model $\rightarrow$ pruned model''. Several state-of-the-art (SOTA) channel pruning methods are compared, including $\ell_1$~\cite{li2017pruning}, SSS~\cite{huang2018data}, GAL~\cite{lin2019towards}, HRank~\cite{lin2020hrank}, SCP~\cite{kang2020operation}, DSA~\cite{ning2020dsa}, SFP~\cite{he2018soft}, FPGM~\cite{he2019filter}, LFPC~\cite{he2020learning}, Rethink~\cite{liu2019rethinking}, ABC~\cite{lin2020channel}, and WM~\cite{howard2017mobilenets}. 
%
%

%
\textbf{VGGNet-16}. 
Table\,\ref{vgg_cifar10} shows the results of pruning 16-layer VGGNet~\cite{simonyan2015very} model, which consists of 13 sequential convolutional layers and 3 fully-connected layers. As can be seen, White-Box yields significantly better top-1 accuracy of 93.47\% compared to the recent SOTA, HRank~\cite{lin2020hrank} of 91.23\%, under similar FLOPs reductions. Moreover, compared to GAL~\cite{lin2019towards} which simply imposes masks upon the outputs of convolutions, our White-Box that considers the internal influence of each channel to the categories, results in a significant reduction on the FLOPs, \emph{i.e.}, 76.64\% \emph{vs.} 39.6\%, while retaining a better top-1 accuracy of 93.47\% \emph{vs.} 92.03\%.

\textbf{ResNet}.
We also evaluate the network pruning performances of various methods on ResNet~\cite{he2016deep},a predominant deep CNN with residual modules, as shown in Table\,\ref{resnet56_cifar10} and Table\,\ref{resnet110_cifar10}. As can be observed, our White-Box increases the performance of original ResNet-56 by 0.28\% and removes around 55.60\% computation burden, while the other methods suffer the accuracy degradation more or less, even reducing less FLOPs. Besides, our White-Box also shows impressive superiority when pruning ResNet-110. With 66.0\% reductions on FLOPs, it still yields 0.55\% performance improvement, surpassing the other methods by a large margin. 

\begin{table}[!t]
\setlength{\tabcolsep}{0.6em}
\centering
\small
\caption{Results for pruning ResNet-110 on CIFAR-10.}
\vspace{-0.5em}
\label{resnet110_cifar10}
\begin{tabular}{ccccc}
\toprule
Model           &Top-1 Acc.& Acc.~$\downarrow$       & FLOPs   $\downarrow$  \\
\midrule
$\ell_1$~\cite{li2017pruning}           &93.55\% $\rightarrow$ 93.30\%     & 0.25\%  &38.7\%   \\
Rethink~\cite{liu2019rethinking} &93.77\% $\rightarrow$ 93.70\%  &0.07\%           & 40.8\%  \\
SFP~\cite{he2018soft} &93.68\% $\rightarrow$ 93.38\% &0.30\%  &40.8\%  \\
GAL~\cite{lin2019towards}     &93.39\% $\rightarrow$ 92.74\%    &0.65\%     &48.5\%  \\
HRank~\cite{lin2020hrank}                       &93.50\% $\rightarrow$ 93.36\%  &0.14\%            & 58.2\%  \\
LFPC~\cite{he2020learning} &93.50\% $\rightarrow$ 93.07\% &0.43\%  &60.3\% \\
ABC~\cite{lin2020channel}      &93.57\% $\rightarrow$ 93.58\% &-0.01\%  &65.0\%  \\
\textbf{White-Box}    &~93.50\% $\rightarrow$ \textbf{94.12\%} &\textbf{-0.62\%} & \textbf{66.0\%} \\
\bottomrule
\end{tabular}

\end{table}

\begin{table}[!t]
\setlength{\tabcolsep}{0.6em}
\small
\centering
\caption{Results for pruning MobileNet-v2 on CIFAR-10.}
\vspace{-0.5em}
\label{mob_cifar10}
\begin{tabular}{cccc}
\toprule
 Model           &Top-1 Acc.& Acc.~$\downarrow$        & FLOPs $\downarrow$  \\

\midrule

WM~\cite{howard2017mobilenets}            &94.47\% $\rightarrow$ 94.02\%    &0.45\%           &27.0\%     \\ 

DCP~\cite{zhuang2018discrimination}            &94.47\% $\rightarrow$ 94.69\%    &-0.22\%          &27.0\%     \\ 

MDP~\cite{guo2020multi} 			&95.02\% $\rightarrow$ 95.14\%    &-0.12\%       &28.7\%     \\ 
\textbf{White-Box}                       &95.02\% $\rightarrow$ \textbf{95.28}\%& ~\textbf{-0.26\%}     &\textbf{29.2\%} \\

\bottomrule
\end{tabular}
\end{table}
%

\begin{table*}[!t]
\caption{\label{imagenet}\normalsize{ Results for pruning ResNet-50 on ILSVRC-2012.}}
\vspace{-0.5em}
\small
\centering
\setlength{\tabcolsep}{0.55em}
\begin{tabular}{ccccccc}
\toprule
Method   &Top-1 Acc. &Top-1 Acc. $\downarrow$ &Top-5 Acc. &Top-5 Acc. $\downarrow$ &FLOPs &FLOPs$\downarrow$ \\ 
\midrule
 SSS-26~\cite{huang2018data}&76.15\% $\rightarrow$ 74.18\%&1.97\% &92.96\% $\rightarrow$ 91.91\% &1.05\%  &2.82G &31.9\%    \\
 CP~\cite{he2017channel}&76.15\% $\rightarrow$ 72.30\%&3.85\% &92.96\% $\rightarrow$ 90.80\%  &2.16\%  &2.73G &34.1\%     \\
   SFP~\cite{he2018soft}  &76.15\% $\rightarrow$ 74.61\%&1.54\% &92.87\% $\rightarrow$ 92.06\% &0.81\%  &2.39G&41.8\%   \\
 GAL~\cite{lin2019towards}  &76.15\% $\rightarrow$ 71.95\%&4.20\% &92.96\% $\rightarrow$ 90.79\% &2.17\%  &2.33G&43.7\%   \\
 SSS-32~\cite{huang2018data}&76.12\% $\rightarrow$ 71.82\%&4.30\% &92.86\% $\rightarrow$ 90.79\% &2.07\%&2.33G &43.7\%  \\
HRank~\cite{lin2020hrank}&76.15\% $\rightarrow$ 75.01\% & 1.14\% &92.96\% $\rightarrow$  92.33\% & 0.63\% &2.30G & 43.9\%  \\
\textbf{White-Box} &76.15\% $\rightarrow$ \textbf{75.32\%}  &\textbf{ 0.83\%} &92.96\% $\rightarrow$ \textbf{92.43\%}  &~\textbf{0.53\%} &\textbf{2.22G}   &  ~\textbf{45.6\% } \\
MetaPruning~\cite{liu2019metapruning} & 76.60\% $\rightarrow$ 75.40\% &1.20\% & -& - & 2.00G & 48.7\% \\
FPGM~\cite{he2019filter} &76.15\% $\rightarrow$ 74.13\%  & 2.02\% &92.96\% $\rightarrow$ 92.87\% & 0.09\% &1.90G &  53.5\%   \\
 RRBP~\cite{zhou2019accelerate} &76.15\% $\rightarrow$ 73.00\%&3.15\% &92.96\% $\rightarrow$ 91.00\% &1.96\% &1.86G&54.5\% \\
 GAL~\cite{lin2019towards} &76.15\% $\rightarrow$ 71.80\%&4.35\% &92.96\% $\rightarrow$ 90.82\% &2.14\% &1.84G&55.6\% \\
 ThiNet~\cite{luo2017thinet} &72.88\% $\rightarrow$ 71.01\%&1.87\% &91.06\% $\rightarrow$ 90.02\%&1.12\%  & 1.71G&58.7\%   \\
 LFPC~\cite{he2020learning} & 76.15\% $\rightarrow$ 74.18\%   & 1.97\% &92.96\% $\rightarrow$ 91.92\%&  1.04\% &1.61G   & 60.8 \%  \\
 HRank~\cite{lin2020hrank} &76.15\% $\rightarrow$ 71.98\%&4.17\% &92.96\% $\rightarrow$ 91.01\% &1.95\% &1.55G &62.6\%   \\
 \textbf{White-Box}   &  76.15\% $\rightarrow$ \textbf{74.21}\%   & ~\textbf{1.94\%}  &~92.96\% $\rightarrow$ \textbf{92.01\% } & \textbf{ 0.95\%}& \textbf{1.50G}     & \textbf{  63.5\% } \\
\bottomrule
\end{tabular}
\end{table*}

%
\textbf{MobileNet-v2}. MobileNet-v2~\cite{sandler2018mobilenetv2} is a prevailing network with a compact design of depth-wise separable convolution. Due to its extremely small computation cost, pruning MobileNet-v2 becomes a particularly challenging task. Nevertheless, compared with the competitors in Tab.\,\ref{mob_cifar10}, White-Box still retains better top-1 accuracy of 95.23\%, while pruning more FLOPs of 29.2\%. 
Further, we plot the performance comparison under different pruning rates of FLOPs in Fig.\,\ref{vary_rate}. To show our advantage, we compare the proposed White-Box with several SOTAs. 
As can be observed, though the pruning rate changes, our White-Box consistently retains a higher top-1 accuracy, which well demonstrates the correctness of exploring the internal CNNs.

\begin{table}[!t]
\setlength{\tabcolsep}{0.8em}
\centering
\small
\caption{\label{speedup}FLOPs, latency and accuracy of White-Box for pruning the ResNet-50. Reported latency is the run-time of the corresponding network on one NVIDIA Tesla V100 GPU with a batch-size of 32.}
\vspace{-0.5em}
\begin{tabular}{ccccc}
\toprule
Method           &FLOPs & Latency&Speedup  &Top-1 Acc.   \\
\midrule
Baseline  & 4.11G  & 1.75ms  & 0.00$\times$  & 76.15\%   \\
White-Box &2.22G  & 1.29ms& 1.35$\times$  & 75.32\%   \\
White-Box  & 1.50G& 1.08ms& 1.62$\times$ & 74.21\%  \\
\bottomrule
\end{tabular}
\end{table}

\subsection{Comparison on ILSVRC-2012}
We further show the results for pruning ResNet-50~\cite{he2016deep} on ILSVRC-2012. In Tab.\,\ref{imagenet}, 
Compared with the SOTAs, White-Box shows the best performance under different pruning rates. By setting $\alpha$ to 0.45, White-Box reduces the FLOPs to around 2.22B while obtaining the top-1 accuracy of 75.32\% and top-5 accuracy of 92.43\%. In contrast, the recent HRank~\cite{lin2020hrank} bears more computation of 2.30 FLOPs and poor top-1 accuracy of 75.01\%  and top-5 accuracy of  92.33\%. Further, we increase $\alpha$ to 0.63 and White-Box shows the least accuracy drops of 2.02\% in  top-1 accuracy  and 1.03\% in top-5 accuracy. With less FLOPs reductions, LFPC~\cite{he2020learning} shows poor top-1 accuracy of 74.18\% and top-5 accuracy of 91.92\%. Tab.\,\ref{speedup} reports the reduction of inference time. our White-Box achieves significant speedups while losing marginal performance. For instance, it obtains 1.62$\times$ GPU speedups with only 1.94\% top-1 accuracy drop, compared with the baseline. 
\subsection{Ablation Study}\label{ablation}

\begin{table}[!t]
\setlength{\tabcolsep}{1em}
\arrayrulecolor{black}
\centering
\small
\caption{Top-1 accuracy comparison for pruning ResNet-56 with varyring configurations on CIFAR-10 under similar FLOPs pruning rate.}
\vspace{-0.5em}
\label{without_classwise}
\begin{tabular}{ccc}
\toprule
Setting           & Top-1 Acc.~$\downarrow$     &  FLOPs   $\downarrow$  \\
\midrule
\textbf{White-Box} & \textbf{-0.28\%} & \textbf{55.6}\% \\  
w/o Class-wise mask& 1.43\% & 53.3\%\\
w/o Soft mask &  2.12\% & 54.7\%\\
w $\ell_1$-norm &0.89\% & 55.2\%\\
\bottomrule
\end{tabular}
\end{table}

\begin{figure}[!t]
\begin{center}
\includegraphics[width=1\linewidth,height=0.6\linewidth]{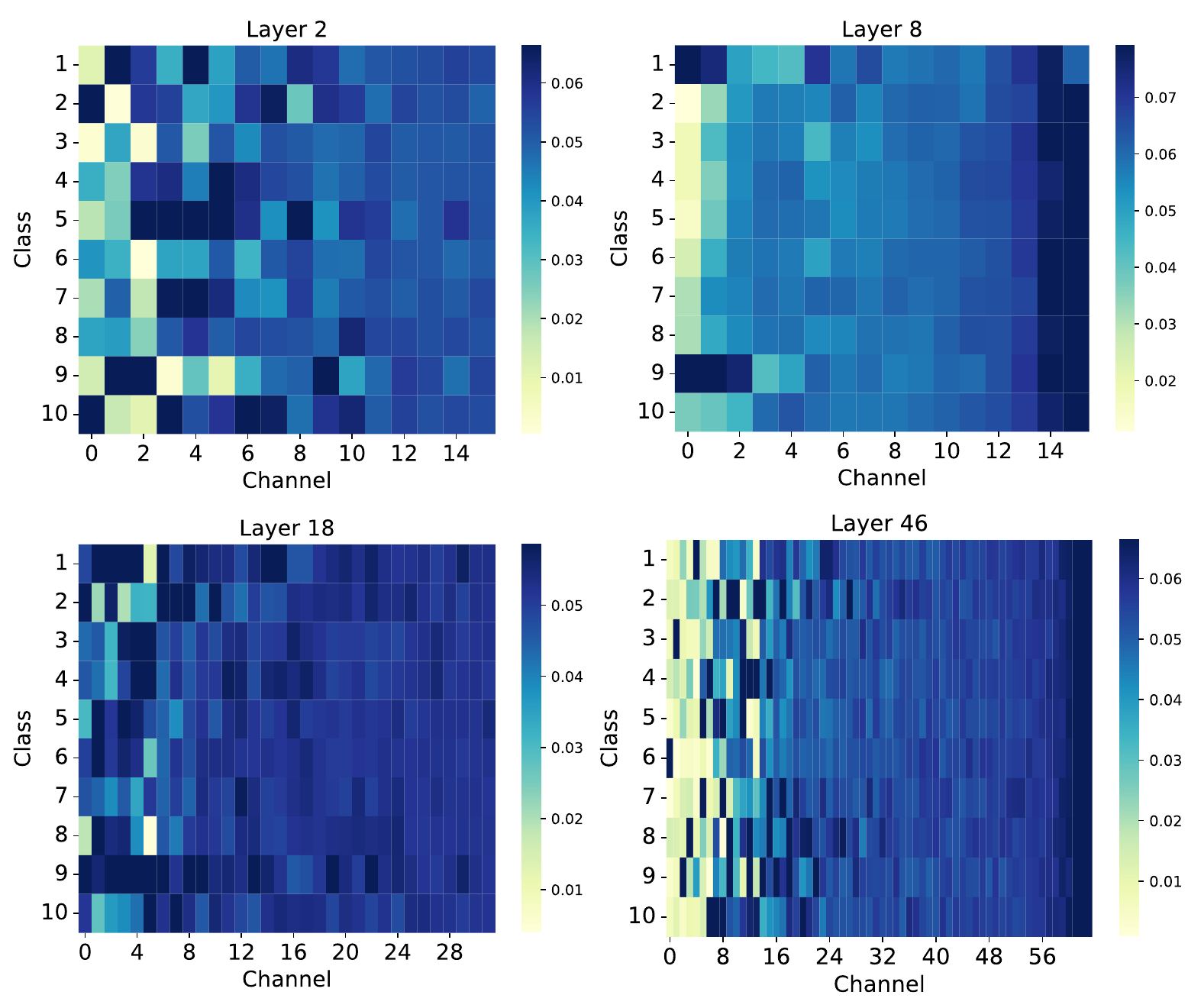}
\end{center}
\setlength{\abovecaptionskip}{0pt}
\setlength{\belowcaptionskip}{0pt}
\vspace{-1.0em}
\caption{\label{mask_ablation}Mask values in varying layers after class-wise training for pruning ResNet-56 on CIFAR-10. The channel axis is sorted by the voting scores.
\vspace{-1.5em}
}
\end{figure}

%
\textbf{Class-wise Mask.}
In this section, we prune ResNet-56 and test its performance on CIFAR-10 as an example to investigate the influences of individual components in our class-wise mask.
We first train each channel with a single mask for all categories of images, denoted as w/o Class-wise mask in Table\,\ref{without_classwise}.
Such a mechanism suffers more accuracy drops as it fails to consider individual channel's discriminating power to recognize different categories as discussed in Sec.\ref{introduction}.
In addition, we conduct experiments without the smooth operation for mask activation, which is referred to as w/o Soft mask.
Table\,\ref{without_classwise} shows that such an implementation leads to the over-fitting problem that the network will converge in one epoch. Thus, the trained mask cannot well contribute to discriminating different categories, leading to an even worse top-1 accuracy than w/o Class-wise mask under a similar FLOPs reduction.
%
%

We also train the class-wise mask with $\ell_1$-norm regularization, denoted as ``w $\ell_1$-norm'' in Table\,\ref{without_classwise}. The $\ell_1$-norm brings worse performance that $\ell_2$-norm, thus demonstrating our motivation of choosing $\ell_2$-norm to measure the class-wise contribution of each channel. Lastly, we show the distribution of the mask value after training over different classes. Visualization in Fig.\ref{mask_ablation} shows that the class-wise mask value changes from channel to channel, which well confirms the motivation of White-Box. Channels with low mask values over all classes will be pruned using the global voting.

\begin{figure}[!t]
\begin{center}
\includegraphics[width=1\linewidth,height=0.45\linewidth]{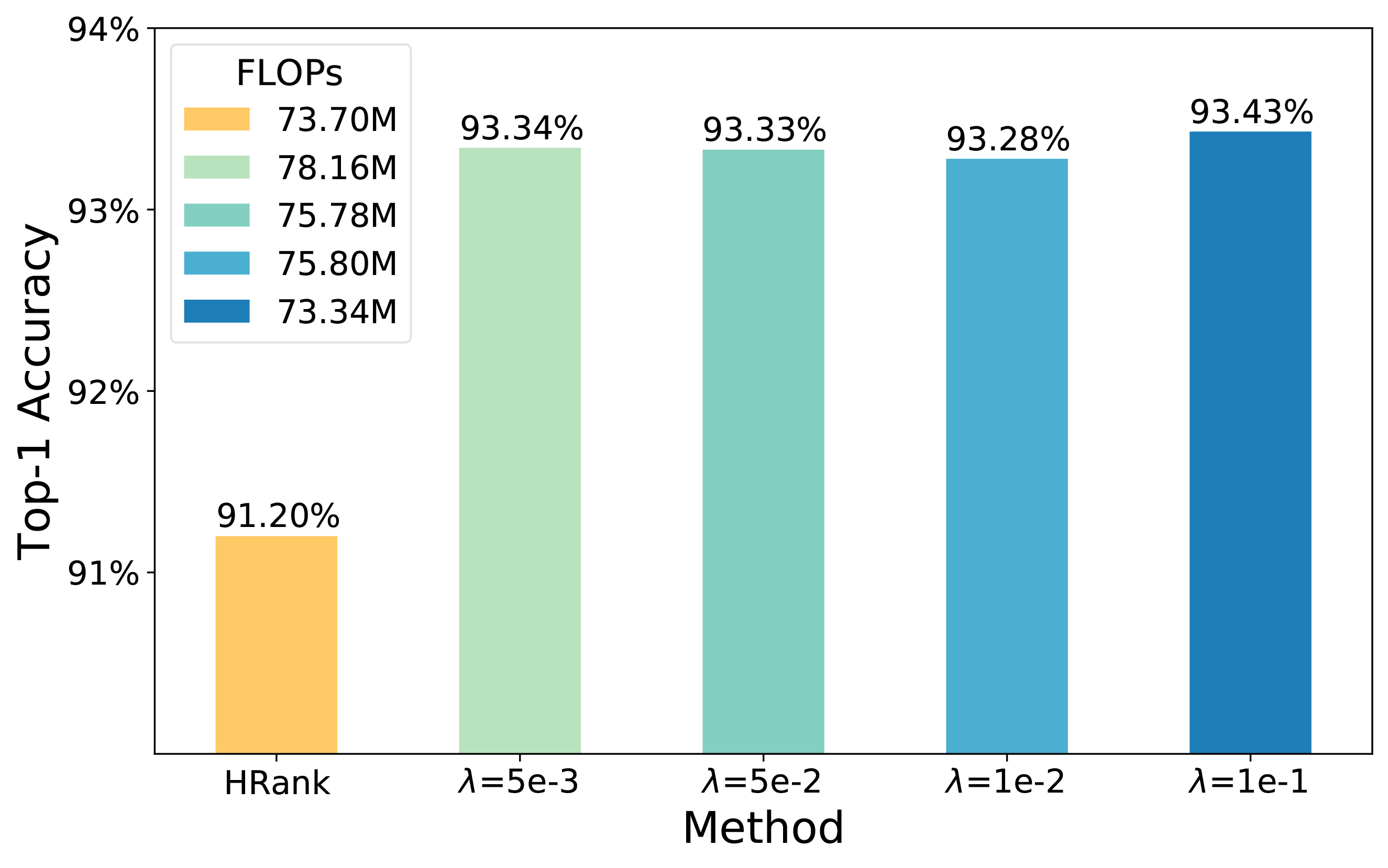}
\end{center}
\setlength{\abovecaptionskip}{0pt}
\setlength{\belowcaptionskip}{0pt}
\vspace{-1.0em}
\caption{\label{vgg_ablation}Top-1 accuracy with different values of $\lambda$ for pruning VGGNet-16 on CIFAR-10 under similar FLOPs pruning rate. Best viewed in color.
\vspace{-1.5em}
}
\end{figure}

\textbf{Sparsity factor.}
We further analyze the impact of the sparsity factor $\lambda$.
We choose to prune VGGNet-16 on the CIFAR-10 with different $\lambda$ under the similar FLOPs pruning rate.
In Fig.\,\ref{vgg_ablation}, with different $\lambda$, all of the pruned networks perform significantly better than the SOTA HRank~\cite{lin2020hrank}.
To explain, our motivation for the class-wise mask training merely falls into observing each channel's contribution to classifying different categories of image, instead of recovering performance.
%
%

%
\section{Conclusion}
Based on visualization and analysis of the deep feature in CNNs, we proposed a new perspective of channel pruning for efficient image classification that one should preserve channels activating discriminative features for more categories in the dataset.
We further carry out channel pruning in a white box by devising a class-wise mask for each channel.
During training, different sub-masks are activated for model inference, \emph{w.r.t.} the current label of input images.
A global voting and a fine-tuning are then performed to obtain the compressed model.
%
%
%



%



\ifCLASSOPTIONcaptionsoff
  \newpage
\fi



%



\bibliographystyle{IEEEtran}
\bibliography{main}

\begin{thebibliography}{10}
\providecommand{\url}[1]{#1}
\csname url@samestyle\endcsname
\providecommand{\newblock}{\relax}
\providecommand{\bibinfo}[2]{#2}
\providecommand{\BIBentrySTDinterwordspacing}{\spaceskip=0pt\relax}
\providecommand{\BIBentryALTinterwordstretchfactor}{4}
\providecommand{\BIBentryALTinterwordspacing}{\spaceskip=\fontdimen2\font plus
\BIBentryALTinterwordstretchfactor\fontdimen3\font minus
  \fontdimen4\font\relax}
\providecommand{\BIBforeignlanguage}[2]{{%
\expandafter\ifx\csname l@#1\endcsname\relax
\typeout{** WARNING: IEEEtran.bst: No hyphenation pattern has been}%
\typeout{** loaded for the language `#1'. Using the pattern for}%
\typeout{** the default language instead.}%
\else
\language=\csname l@#1\endcsname
\fi
#2}}
\providecommand{\BIBdecl}{\relax}
\BIBdecl

\bibitem{simonyan2015very}
K.~Simonyan and A.~Zisserman, ``Very deep convolutional networks for
  large-scale image recognition,'' in \emph{International Conference on
  Learning Representations (ICLR)}, 2015.

\bibitem{he2016deep}
K.~He, X.~Zhang, S.~Ren, and J.~Sun, ``Deep residual learning for image
  recognition,'' in \emph{IEEE Conference on Computer Vision and Pattern
  Recognition (CVPR)}, 2016, pp. 770--778.

\bibitem{han2015learning}
S.~Han, J.~Pool, J.~Tran, and W.~Dally, ``Learning both weights and connections
  for efficient neural network,'' in \emph{Advances in Neural Information
  Processing Systems (NeurIPS)}, 2015, pp. 1135--1143.

\bibitem{ding2019global}
X.~Ding, X.~Zhou, Y.~Guo, J.~Han, J.~Liu \emph{et~al.}, ``Global sparse
  momentum sgd for pruning very deep neural networks,'' in \emph{Advances in
  Neural Information Processing Systems (NeurIPS)}, 2019, pp. 6382--6394.

\bibitem{hubara2016binarized}
I.~Hubara, M.~Courbariaux, D.~Soudry, R.~El-Yaniv, and Y.~Bengio, ``Binarized
  neural networks,'' in \emph{Advances in Neural Information Processing Systems
  (NeurIPS)}, 2016, pp. 4107--4115.

\bibitem{liu2020bi}
Z.~Liu, W.~Luo, B.~Wu, X.~Yang, W.~Liu, and K.-T. Cheng, ``Bi-real net:
  Binarizing deep network towards real-network performance,''
  \emph{International Journal of Computer Vision (IJCV)}, vol. 128, no.~1, pp.
  202--219, 2020.

\bibitem{peng2018extreme}
B.~Peng, W.~Tan, Z.~Li, S.~Zhang, D.~Xie, and S.~Pu, ``Extreme network
  compression via filter group approximation,'' in \emph{European Conference on
  Computer Vision (ECCV)}, 2018, pp. 300--316.

\bibitem{hayashi2019exploring}
K.~Hayashi, T.~Yamaguchi, Y.~Sugawara, and S.-i. Maeda, ``Exploring unexplored
  tensor network decompositions for convolutional neural networks,'' in
  \emph{Advances in Neural Information Processing Systems (NeurIPS)}, 2019, pp.
  5552--5562.

\bibitem{romero2014fitnets}
A.~Romero, N.~Ballas, S.~E. Kahou, A.~Chassang, C.~Gatta, and Y.~Bengio,
  ``Fitnets: Hints for thin deep nets,'' \emph{arXiv preprint arXiv:1412.6550},
  2014.

\bibitem{hinton2015distilling}
G.~Hinton, O.~Vinyals, and J.~Dean, ``Distilling the knowledge in a neural
  network,'' \emph{arXiv preprint arXiv:1503.02531}, 2015.

\bibitem{li2017pruning}
H.~Li, A.~Kadav, I.~Durdanovic, H.~Samet, and H.~P. Graf, ``Pruning filters for
  efficient convnets,'' in \emph{International Conference on Learning
  Representations (ICLR)}, 2017.

\bibitem{he2019filter}
Y.~He, P.~Liu, Z.~Wang, Z.~Hu, and Y.~Yang, ``Filter pruning via geometric
  median for deep convolutional neural networks acceleration,'' in \emph{IEEE
  Conference on Computer Vision and Pattern Recognition (CVPR)}, 2019, pp.
  4340--4349.

\bibitem{hu2016network}
H.~Hu, R.~Peng, Y.-W. Tai, and C.-K. Tang, ``Network trimming: A data-driven
  neuron pruning approach towards efficient deep architectures,'' \emph{arXiv
  preprint arXiv:1607.03250}, 2016.

\bibitem{lin2020hrank}
M.~Lin, R.~Ji, Y.~Wang, Y.~Zhang, B.~Zhang, Y.~Tian, and L.~Shao, ``Hrank:
  Filter pruning using high-rank feature map,'' in \emph{IEEE Conference on
  Computer Vision and Pattern Recognition (CVPR)}, 2020, pp. 1529--1538.

\bibitem{guo2020channel}
J.~Guo, W.~Ouyang, and D.~Xu, ``Channel pruning guided by classification loss
  and feature importance,'' in \emph{AAAI Conference on Artificial Intelligence
  (AAAI)}, 2020, pp. 10\,885--10\,892.

\bibitem{liu2017learning}
Z.~Liu, J.~Li, Z.~Shen, G.~Huang, S.~Yan, and C.~Zhang, ``Learning efficient
  convolutional networks through network slimming,'' in \emph{IEEE
  International Conference on Computer Vision (ICCV)}, 2017, pp. 2736--2744.

\bibitem{luo2020autopruner}
J.-H. Luo and J.~Wu, ``Autopruner: An end-to-end trainable filter pruning
  method for efficient deep model inference,'' \emph{Pattern Recognition (PR)},
  p. 107461, 2020.

\bibitem{ding2019centripetal}
X.~Ding, G.~Ding, Y.~Guo, and J.~Han, ``Centripetal sgd for pruning very deep
  convolutional networks with complicated structure,'' in \emph{Proceedings of
  the IEEE/CVF Conference on Computer Vision and Pattern Recognition}, 2019,
  pp. 4943--4953.

\bibitem{ding2020lossless}
X.~Ding, T.~Hao, J.~Liu, J.~Han, Y.~Guo, and G.~Ding, ``Lossless cnn channel
  pruning via gradient resetting and convolutional re-parameterization,''
  \emph{arXiv preprint arXiv:2007.03260}, 2020.

\bibitem{he2018amc}
Y.~He, J.~Lin, Z.~Liu, H.~Wang, L.-J. Li, and S.~Han, ``Amc: Automl for model
  compression and acceleration on mobile devices,'' in \emph{European
  Conference on Computer Vision (ECCV)}, 2018, pp. 784--800.

\bibitem{yang2018netadapt}
T.-J. Yang, A.~Howard, B.~Chen, X.~Zhang, A.~Go, M.~Sandler, V.~Sze, and
  H.~Adam, ``Netadapt: Platform-aware neural network adaptation for mobile
  applications,'' in \emph{European Conference on Computer Vision (ECCV)},
  2018, pp. 285--300.

\bibitem{liu2019metapruning}
Z.~Liu, H.~Mu, X.~Zhang, Z.~Guo, X.~Yang, T.~K.-T. Cheng, and J.~Sun,
  ``Metapruning: Meta learning for automatic neural network channel pruning,''
  in \emph{IEEE International Conference on Computer Vision (ICCV)}, 2019, pp.
  3296--3305.

\bibitem{liu2021joint}
Z.~Liu, X.~Zhang, Z.~Shen, Y.~Wei, K.-T. Cheng, and J.~Sun, ``Joint
  multi-dimension pruning via numerical gradient update,'' \emph{IEEE
  Transactions on Image Processing (TIP)}, vol.~30, pp. 8034--8045, 2021.

\bibitem{ding2019approximated}
X.~Ding, G.~Ding, Y.~Guo, J.~Han, and C.~Yan, ``Approximated oracle filter
  pruning for destructive cnn width optimization,'' in \emph{International
  Conference on Machine Learning (ICML)}.\hskip 1em plus 0.5em minus
  0.4em\relax PMLR, 2019, pp. 1607--1616.

\bibitem{wu2017interpretable}
T.~Wu, X.~Li, X.~Song, W.~Sun, L.~Dong, and B.~Li, ``Interpretable r-cnn,''
  \emph{arXiv preprint arXiv:1711.05226}, vol.~2, 2017.

\bibitem{yosinski2015understanding}
J.~Yosinski, J.~Clune, A.~Nguyen, T.~Fuchs, and H.~Lipson, ``Understanding
  neural networks through deep visualization,'' in \emph{Deep Learning
  Workshop, International Conference on Machine Learning (ICML)}, 2015.

\bibitem{zeiler2014visualizing}
M.~D. Zeiler and R.~Fergus, ``Visualizing and understanding convolutional
  networks,'' in \emph{European conference on computer vision (ECCV)}.\hskip
  1em plus 0.5em minus 0.4em\relax Springer, 2014, pp. 818--833.

\bibitem{zhang2018interpretable}
Q.~Zhang, Y.~Nian~Wu, and S.-C. Zhu, ``Interpretable convolutional neural
  networks,'' in \emph{Proceedings of the IEEE Conference on Computer Vision
  and Pattern Recognition (CVPR)}, 2018, pp. 8827--8836.

\bibitem{zhou2014object}
B.~Zhou, A.~Khosla, {\`{A}}.~Lapedriza, A.~Oliva, and A.~Torralba, ``Object
  detectors emerge in deep scene cnns,'' in \emph{International Conference on
  Learning Representations (ICLR)}, 2015.

\bibitem{shi2020sasl}
J.~Shi, J.~Xu, K.~Tasaka, and Z.~Chen, ``Sasl: saliency-adaptive sparsity
  learning for neural network acceleration,'' \emph{IEEE Transactions on
  Circuits and Systems for Video Technology}, vol.~31, no.~5, pp. 2008--2019,
  2020.

\bibitem{he2017channel}
Y.~He, X.~Zhang, and J.~Sun, ``Channel pruning for accelerating very deep
  neural networks,'' in \emph{IEEE International Conference on Computer Vision
  (ICCV)}, 2017, pp. 1389--1397.

\bibitem{guo2020model}
J.~Guo, W.~Zhang, W.~Ouyang, and D.~Xu, ``Model compression using progressive
  channel pruning,'' \emph{IEEE Transactions on Circuits and Systems for Video
  Technology}, vol.~31, no.~3, pp. 1114--1124, 2020.

\bibitem{yu2020antidote}
F.~Yu, C.~Liu, D.~Wang, Y.~Wang, and X.~Chen, ``Antidote: attention-based
  dynamic optimization for neural network runtime efficiency,'' in \emph{2020
  Design, Automation \& Test in Europe Conference \& Exhibition (DATE)}.\hskip
  1em plus 0.5em minus 0.4em\relax IEEE, 2020, pp. 951--956.

\bibitem{yamamoto2018pcas}
K.~Yamamoto and K.~Maeno, ``Pcas: Pruning channels with attention statistics
  for deep network compression,'' in \emph{British Machine Vision Conference
  (BMVC)}, 2018.

\bibitem{wang2019convolutional}
X.-J. Wang, W.~Yao, and H.~Fu, ``A convolutional neural network pruning method
  based on attention mechanism,'' in \emph{SEKE}, 2019, pp. 343--452.

\bibitem{huang2018data}
Z.~Huang and N.~Wang, ``Data-driven sparse structure selection for deep neural
  networks,'' in \emph{European Conference on Computer Vision (ECCV)}, 2018,
  pp. 304--320.

\bibitem{xiao2019autoprune}
X.~Xiao, Z.~Wang, and S.~Rajasekaran, ``Autoprune: Automatic network pruning by
  regularizing auxiliary parameters,'' in \emph{Advances in Neural Information
  Processing Systems (NeurIPS)}, 2019, pp. 13\,681--13\,691.

\bibitem{chen2020dynamical}
Z.~Chen, T.-B. Xu, C.~Du, C.-L. Liu, and H.~He, ``Dynamical channel pruning by
  conditional accuracy change for deep neural networks,'' \emph{IEEE
  transactions on neural networks and learning systems}, vol.~32, no.~2, pp.
  799--813, 2020.

\bibitem{tang2021manifold}
Y.~Tang, Y.~Wang, Y.~Xu, Y.~Deng, C.~Xu, D.~Tao, and C.~Xu, ``Manifold
  regularized dynamic network pruning,'' in \emph{Proceedings of the IEEE/CVF
  Conference on Computer Vision and Pattern Recognition}, 2021, pp. 5018--5028.

\bibitem{szegedy2016rethinking}
C.~Szegedy, V.~Vanhoucke, S.~Ioffe, J.~Shlens, and Z.~Wojna, ``Rethinking the
  inception architecture for computer vision,'' in \emph{IEEE Conference on
  Computer Vision and Pattern Recognition (CVPR)}, 2016, pp. 2818--2826.

\bibitem{liu2019rethinking}
Z.~Liu, M.~Sun, T.~Zhou, G.~Huang, and T.~Darrell, ``Rethinking the value of
  network pruning,'' in \emph{International Conference on Learning
  Representations (ICLR)}, 2019.

\bibitem{li2020eagleeye}
B.~Li, B.~Wu, J.~Su, G.~Wang, and L.~Lin, ``Eagleeye: Fast sub-net evaluation
  for efficient neural network pruning,'' \emph{arXiv preprint
  arXiv:2007.02491}, 2020.

\bibitem{krizhevsky2009learning}
A.~Krizhevsky, G.~Hinton \emph{et~al.}, ``Learning multiple layers of features
  from tiny images,'' 2009.

\bibitem{russakovsky2015imagenet}
O.~Russakovsky, J.~Deng, H.~Su, J.~Krause, S.~Satheesh, S.~Ma, Z.~Huang,
  A.~Karpathy, A.~Khosla, M.~Bernstein \emph{et~al.}, ``Imagenet large scale
  visual recognition challenge,'' \emph{International Journal of Computer
  Vision (IJCV)}, vol. 115, no.~3, pp. 211--252, 2015.

\bibitem{sandler2018mobilenetv2}
M.~Sandler, A.~Howard, M.~Zhu, A.~Zhmoginov, and L.-C. Chen, ``Mobilenetv2:
  Inverted residuals and linear bottlenecks,'' in \emph{IEEE Conference on
  Computer Vision and Pattern Recognition (CVPR)}, 2018, pp. 4510--4520.

\bibitem{pytorch2015}
A.~Paszke, S.~Gross, F.~Massa, A.~Lerer, J.~Bradbury, G.~Chanan, T.~Killeen,
  Z.~Lin, N.~Gimelshein, L.~Antiga \emph{et~al.}, ``Pytorch: An imperative
  style, high-performance deep learning library,'' in \emph{Advances in Neural
  Information Processing Systems (NeurIPS)}, 2019, pp. 8026--8037.

\bibitem{lin2019towards}
S.~Lin, R.~Ji, C.~Yan, B.~Zhang, L.~Cao, Q.~Ye, F.~Huang, and D.~Doermann,
  ``Towards optimal structured cnn pruning via generative adversarial
  learning,'' in \emph{IEEE Conference on Computer Vision and Pattern
  Recognition (CVPR)}, 2019, pp. 2790--2799.

\bibitem{kang2020operation}
M.~Kang and B.~Han, ``Operation-aware soft channel pruning using differentiable
  masks,'' in \emph{International Conference on Machine Learning (ICML)}, 2020.

\bibitem{he2018soft}
Y.~He, G.~Kang, X.~Dong, Y.~Fu, and Y.~Yang, ``Soft filter pruning for
  accelerating deep convolutional neural networks,'' in \emph{International
  Joint Conference on Artificial Intelligence (IJCAI)}, 2018, pp. 2234--2240.

\bibitem{he2020learning}
Y.~He, Y.~Ding, P.~Liu, L.~Zhu, H.~Zhang, and Y.~Yang, ``Learning filter
  pruning criteria for deep convolutional neural networks acceleration,'' in
  \emph{IEEE Conference on Computer Vision and Pattern Recognition (CVPR)},
  2020, pp. 2009--2018.

\bibitem{ning2020dsa}
X.~Ning, T.~Zhao, W.~Li, P.~Lei, Y.~Wang, and H.~Yang, ``Dsa: More efficient
  budgeted pruning via differentiable sparsity allocation,'' \emph{arXiv
  preprint arXiv:2004.02164}, 2020.

\bibitem{lin2020channel}
M.~Lin, R.~Ji, Y.~Zhang, B.~Zhang, Y.~Wu, and Y.~Tian, ``Channel pruning via
  automatic structure search,'' in \emph{International Joint Conference on
  Artificial Intelligence (IJCAI)}, 2020, pp. 673--679.

\bibitem{howard2017mobilenets}
A.~G. Howard, M.~Zhu, B.~Chen, D.~Kalenichenko, W.~Wang, T.~Weyand,
  M.~Andreetto, and H.~Adam, ``Mobilenets: Efficient convolutional neural
  networks for mobile vision applications,'' \emph{arXiv preprint
  arXiv:1704.04861}, 2017.

\bibitem{zhuang2018discrimination}
Z.~Zhuang, M.~Tan, B.~Zhuang, J.~Liu, Y.~Guo, Q.~Wu, J.~Huang, and J.~Zhu,
  ``Discrimination-aware channel pruning for deep neural networks,'' in
  \emph{Advances in Neural Information Processing Systems (NeurIPS)}, 2018, pp.
  875--886.

\bibitem{guo2020multi}
J.~Guo, W.~Ouyang, and D.~Xu, ``Multi-dimensional pruning: A unified framework
  for model compression,'' in \emph{IEEE Conference on Computer Vision and
  Pattern Recognition (CVPR)}, 2020, pp. 1508--1517.

\bibitem{zhou2019accelerate}
Y.~Zhou, Y.~Zhang, Y.~Wang, and Q.~Tian, ``Accelerate cnn via recursive
  bayesian pruning,'' in \emph{IEEE International Conference on Computer Vision
  (CVPR)}, 2019, pp. 3306--3315.

\bibitem{luo2017thinet}
J.-H. Luo, J.~Wu, and W.~Lin, ``Thinet: A filter level pruning method for deep
  neural network compression,'' in \emph{IEEE International Conference on
  Computer Vision (ICCV)}, 2017, pp. 5058--5066.

\end{thebibliography}

\ifCLASSOPTIONcaptionsoff
  \newpage
\fi

%

%


\begin{IEEEbiography}[{\includegraphics[width=1in,height=1.25in,clip,keepaspectratio]{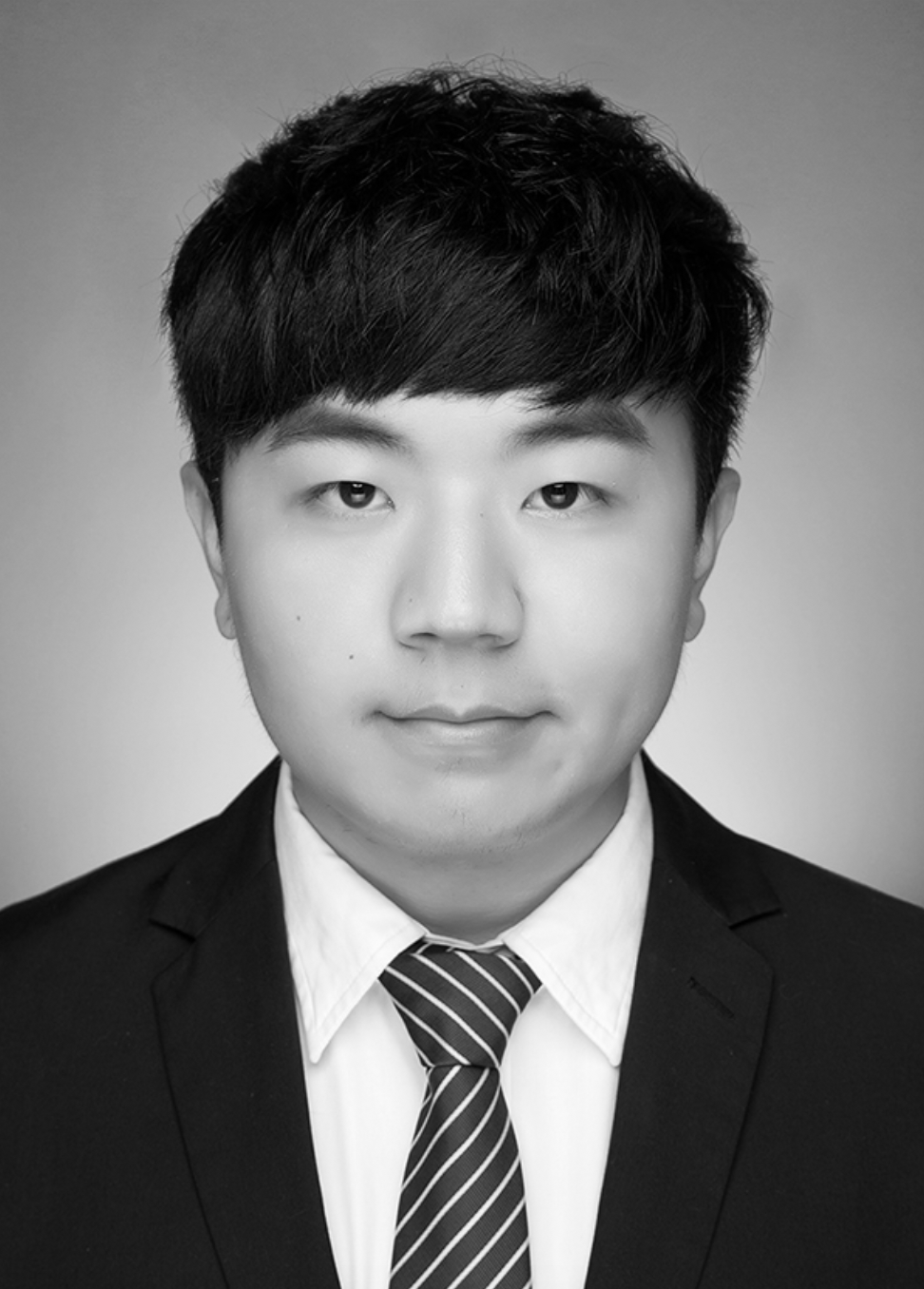}}]{Yuxin Zhang} received the B.E. degree in Computer Science, School of Informatics, Xiamen University, Xiamen, China, in 2020.
He is currently pursuing the B.S. degree with Xiamen University, China. His research interests include computer vision and neural network compression \& acceleration.
\end{IEEEbiography}

\begin{IEEEbiography}[{\includegraphics[width=1in,height=1.25in,clip,keepaspectratio]{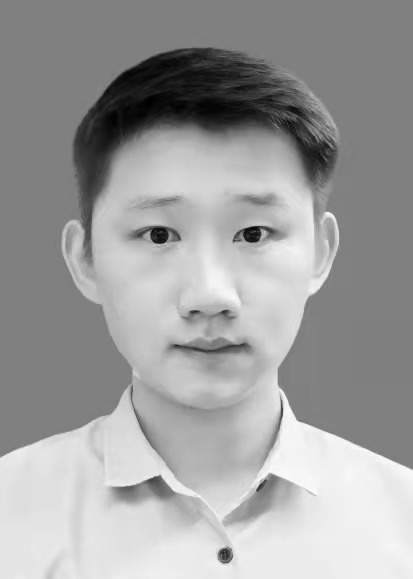}}]{Mingbao Lin} is currently pursuing the Ph.D degree with Xiamen University, China. He has published over ten papers as the first author in international journals and conferences, including IEEE TPAMI, IJCV, IEEE TIP, IEEE TNNLS, IEEE CVPR, NeuriPS, AAAI, IJCAI, ACM MM and so on. His current research interest includes network compression \& acceleration, and information retrieval.
\end{IEEEbiography}

\begin{IEEEbiography}[{\includegraphics[width=1in,height=1.25in,clip,keepaspectratio]{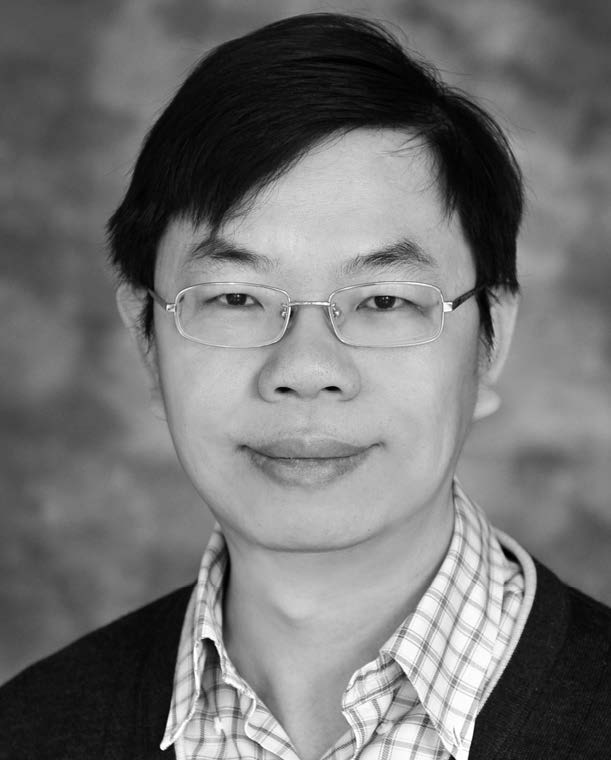}}]{Chia-Wen Lin} (Fellow, IEEE) received the Ph.D degree in electrical engineering from National Tsing Hua University (NTHU), Hsinchu, Taiwan, in 2000.
He is currently Professor with the Department of Electrical Engineering and the Institute of Communications Engineering, NTHU. He is also Deputy Director of the AI Research Center of NTHU. He was with the Department of Computer Science and Information Engineering, National Chung Cheng University, Taiwan, during 2000--2007. Prior to joining academia, he worked for the Information and Communications Research Laboratories, Industrial Technology Research Institute, Hsinchu, Taiwan, during 1992--2000. His research interests include image and video processing, computer vision, and video networking.

Dr. Lin served as  Distinguished Lecturer of IEEE Circuits and Systems Society from 2018 to 2019, a Steering Committee member of \textsc{IEEE Transactions on Multimedia} from 2014 to 2015, and the Chair of the Multimedia Systems and Applications Technical Committee of the IEEE Circuits and Systems Society from 2013 to 2015.  His articles received the Best Paper Award of IEEE VCIP 2015 and the Young Investigator Award of VCIP 2005. He received Outstanding Electrical Professor Award presented by Chinese Institute of Electrical Engineering in 2019, and Young Investigator Award presented by Ministry of Science and Technology, Taiwan, in 2006. He is also the Chair of the Steering Committee of IEEE ICME.  He has served as a Technical Program Co-Chair for IEEE ICME 2010, and a General Co-Chair for IEEE VCIP 2018, and a Technical Program Co-Chair for IEEE ICIP 2019. He has served as an Associate Editor of \textsc{IEEE Transactions on Image Processing}, \textsc{IEEE Transactions on Circuits and Systems for Video Technology}, \textsc{IEEE Transactions on Multimedia}, \textsc{IEEE Multimedia}, and \textit{Journal of Visual Communication and Image Representation}. 
\end{IEEEbiography}

\begin{IEEEbiography} [{\includegraphics[width=1in,height=1.25in,clip,keepaspectratio]{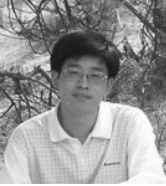}}]{Jie Chen}
received his M.S. and Ph.D. degrees from Harbin Institute of Technology, China, in 2002 and 2007, respectively. He joined Peng Cheng Labora- tory, China since 2018. He has been a senior researcher in the Center for Machine Vision and Signal Analysis at the University of Oulu, Finland since 2007. In 2012 and 2015, he visited the Computer Vision Laboratory at the University of Maryland and School of Electrical and Computer Engineering at the Duke University respectively. Dr. Chen was a cochair of International Workshops at ACCV, CVPR,and ICCV. He was a guest editor of special issues for IEEE TPAMI and IJCV. His research interests include pattern recognition, computer vision, machine learning, dynamic texture, deep learning, and medical image analysis.
\end{IEEEbiography}

\begin{IEEEbiography}[{\includegraphics[width=1in,height=1.25in,clip,keepaspectratio]{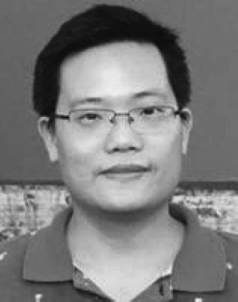}}]{Yongjian Wu} received the master’s degree in computer science from Wuhan University, Wuhan, China, in 2008. 

He is currently the Expert Researcher and the Director of the Youtu Laboratory, Tencent Co., Ltd., Shanghai, China. His research interests include face recognition, image understanding, and large-scale data processing.
\end{IEEEbiography}

\begin{IEEEbiography}[{\includegraphics[width=1in,height=1.25in,clip,keepaspectratio]{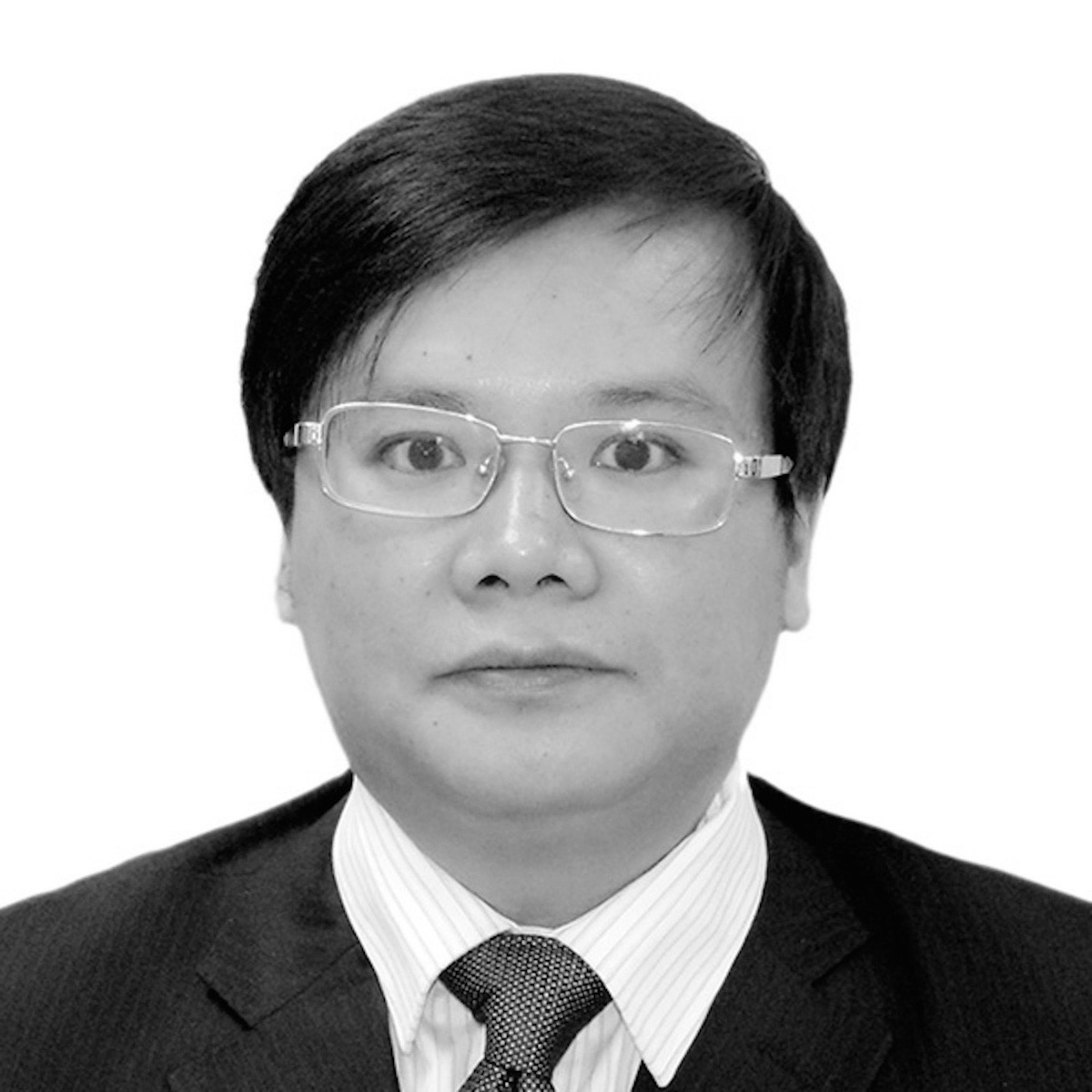}}]{Yonghong Tian} (Senior Member, IEEE) is currently a Boya Distinguished Professor with the Department of Computer Science and Technology, Peking University, Beijing, China, and is also the Deputy Director of the Artificial Intelligence Research Center, PengCheng Laboratory, Shenzhen, China. He has authored or coauthored over 200 technical articles in refereed journals, such as the IEEE TRANSACTIONS ON PATTERN ANALYS IS AND MACHINE INTELLIGENCE (TPAMI)/TRANSACTIONS ON NEURAL NETWORKS AND LEARNING SYSTEMS (TNNLS)/TRANSACTIONS ON IMAGE PROCESSING (TIP)/TRANSACTIONS ON MULTIMEDIA (TMM)/TRANSACTIONS ON CIRCUITS AND SYSTEMS FOR VIDEO TECHNOLOGY (TCSVT)/TRANSACTIONS ON KNOWLEDGE AND DATA ENGINEERING (TKDE)/TRANSACTIONS ON PARALLEL AND DISTRIBUTED SYSTEMS (TPDS), the ACM Computing Surveys (CSUR)/Transactions on Information Systems (TOIS)/Transactions on Multimedia Computing, Communications, and Applications (TOMM), and conferences, such as Neural Information Processing Systems (NeuriPS)/the IEEE Conference on Computer Vision and Pattern Recognition (CVPR)/the International Conference on Computer Vision (ICCV)/AAAI/the ACM International Conference on Multimedia /the International World Wide Web Conference. His research interests include neuromorphic vision, brain-inspired computation, and multimedia big data. 

Prof. Tian is a Senior Member of the CIE and CCF and a member of the ACM. He was a recipient of the Chinese National Science Foundation for Distinguished Young Scholars in 2018, two National Science and Technology Awards, and three ministerial-level awards in China, and obtained the 2015 EURASIP Best Paper Award for the Journal on Image and Video Processing, and the Best Paper Award of the IEEE BigMM 2018. He has been an Associate Editor of the IEEE TRANSACTIONS ON CIRCUITS AND SYSTEMS FOR VIDEO TECHNOLOGY (TCSVT) since 2018, the IEEE Multimedia Magazine since 2018.1, and the IEEE ACCESS since January 2017, and was an Associate Editor of the IEEE TRANSACTIONS ON MULTIMEDIA (TMM) from August 2014 to August 2018. He co-initiated the IEEE International Conference on Multimedia Big Data (BigMM) and served as the TPC Co-Chair for BigMM 2015, and also served as the Technical Program Co-Chair for the IEEE International Conference on Multimedia and Expo (ICME) 2015, the IEEE ISM 2015, and the IEEE International Conference on Multimedia Information Processing and Retrieval (MIPR) 2018/2019, and the General Co-Chair of the IEEE MIPR 2020 and ICME 2021. He has been a Steering Member of the IEEE ICME since 2018 and the IEEE BigMM since 2015, and is a TPC Member of more than ten conferences, such as the Conference on Computer Vision and Pattern Recognition (CVPR), the International Conference on Computer Vision (ICCV), the ACM Transactions on Knowledge Discovery from Data, AAAI, the ACM Multimedia (MM), and the European Conference on Computer Vision (ECCV).
\end{IEEEbiography}

\begin{IEEEbiography}[{\includegraphics[width=1in,height=1.25in,clip,keepaspectratio]{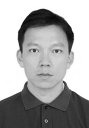}}]{Rongrong Ji}
(Senior Member, IEEE) is a Nanqiang Distinguished Professor at Xiamen University, the Deputy Director of the Office of Science and Technology at Xiamen University, and the Director of Media Analytics and Computing Lab. He was awarded as the National Science Foundation for Excellent Young Scholars (2014), the National Ten Thousand Plan for Young Top Talents (2017), and the National Science Foundation for Distinguished Young Scholars (2020). His research falls in the field of computer vision, multimedia analysis, and machine learning. He has published 50+ papers in ACM/IEEE Transactions, including TPAMI and IJCV, and 100+ full papers on top-tier conferences, such as CVPR and NeurIPS. His publications have got over 10K citations in Google Scholar. He was the recipient of the Best Paper Award of ACM Multimedia 2011. He has served as Area Chairs in top-tier conferences such as CVPR and ACM Multimedia. He is also an Advisory Member for Artificial Intelligence Construction in the Electronic Information Education Committee of the National Ministry of Education.
\end{IEEEbiography}




\end{document}